\documentclass{llncs}

\usepackage[english]{babel}

\usepackage[letterpaper,top=2cm,bottom=2cm,left=3cm,right=3cm,marginparwidth=1.75cm]{geometry}

\usepackage{cite}
\usepackage{amssymb,bm}
\usepackage{amsmath}
\usepackage{graphicx}
\usepackage[colorlinks=true, allcolors=blue]{hyperref}
\usepackage{bbding}
\usepackage{booktabs}
\usepackage{orcidlink}
\usepackage{multirow}
\usepackage{multicol}

\usepackage{marvosym}

\begin{document}

\title{InstEditSeg: Instruction-Driven Image Editing for Polyp and Skin Lesion Segmentation}

\author{Ziquan Liu\orcidlink{0009-0007-3587-1542} \and
Zhewei Zhu\orcidlink{0009-0007-4661-0516} \and
Xuyang Shi\textsuperscript{\Letter} \orcidlink{0000-0002-9473-8692}}

\institute{School of Information and Control Engineering, Southwest University of Science and Technology, Mianyang 621010, China \\
\email{\{ziquanliu, zhuzw\}@mails.swust.edu.cn}, \email{xuyangshi@swust.edu.cn}}

\maketitle

\begin{abstract}
Accurate segmentation of polyps and skin lesions is pivotal for clinical diagnosis, yet existing methods struggle with low contrast, ambiguous boundaries, and cross-domain distribution discrepancies. Discriminative networks and most diffusion-based segmentation approaches predict standalone binary masks, leaving the visual priors of large-scale pretrained generative models largely unexploited. We propose InstEditSeg, a unified generative framework that reformulates medical segmentation as an instruction-driven image editing problem. Instead of emitting a mask, the model renders a color-coded overlay on the original image, conditioned on a textual instruction, so that the edited output aligns with the natural image distribution learned by latent diffusion models and mitigates the domain gap between natural and medical imagery. To recover fine anatomical structures, we introduce DINOv3 as an auxiliary visual encoder and a DINO Feature Guidance Block that builds a multi-scale feature pyramid. The pyramid is fused into the diffusion U-Net by channel concatenation and zero-initialized convolution so that hierarchical discriminative priors can be injected without perturbing the pretrained weights. A dual-branch classifier-free guidance strategy requiring only two forward passes per denoising step reduces inference cost. On polyp and skin lesion benchmarks the framework achieves accuracy competitive with strong discriminative baselines, and it further demonstrates concrete advantages of the generative formulation: notably better cross-domain generalization on unseen data, more complete multi-lesion segmentation, instruction-conditioned task control, and sampling flexibility. We also analyze the strengths and limitations of the paradigm, including its color sensitivity and unsupported attribute-conditioned selection. Code is available at: \href{https://github.com/wincharm001/InstEditSeg}{https://github.com/wincharm001/InstEditSeg}
\end{abstract}

\section{Introduction}
\label{sec:introduction}
Colorectal cancer (CRC) is a leading cause of cancer-related mortality worldwide, and most cases originate from adenomatous polyps. Accurate polyp segmentation during colonoscopy is therefore crucial for early diagnosis, but it remains challenging due to low contrast, ambiguous boundaries, diverse morphology, and imaging artifacts such as uneven illumination and specular reflections, which hinder both accuracy and generalization.

Similarly, accurate segmentation of skin lesions in dermoscopic images is essential for the early diagnosis of melanoma \cite{gutman2016skin}, yet it faces comparable challenges, including fuzzy lesion borders, low contrast, and high inter-patient variability.

Deep learning has improved medical segmentation substantially. U-Net \cite{ronneberger2015u} became the de facto standard, while foundation models such as the Segment Anything Model (SAM) \cite{kirillov2023segany} show strong zero-shot capabilities. Nevertheless, SAM is trained on natural images and suffers substantial domain gaps on medical data, especially for small lesions and blurry boundaries.

Diffusion probabilistic models \cite{ho2020denoising} offer an alternative paradigm. SegDiff \cite{amit2021segdiff} casts segmentation as iterative denoising, and MedSegDiff and MedSegDiff-V2 \cite{wu2024medsegdiff, wu2024medsegdiffv2} extend this to medical tasks. These approaches benefit from generative capabilities but still predict task-specific binary masks, making them specialized diffusion models rather than a unified generative framework.

Instruction-driven image editing \cite{brooks2023instructpix2pix} and unified visual generation models \cite{visionbanana2026, sensenova2026sensenovavision} show that detection, segmentation, and depth can be formulated as conditional generation. Using textual instructions, these models unify tasks within a single diffusion network and can produce color-coded segmentation directly, bridging visual understanding and generation. Applying them to medical images, however, faces two limitations: the substantial distribution gap between natural and medical images, and the limited spatial guidance of text alone for fine anatomical structures.

\begin{figure}
\centering
\includegraphics[width=0.4\linewidth]{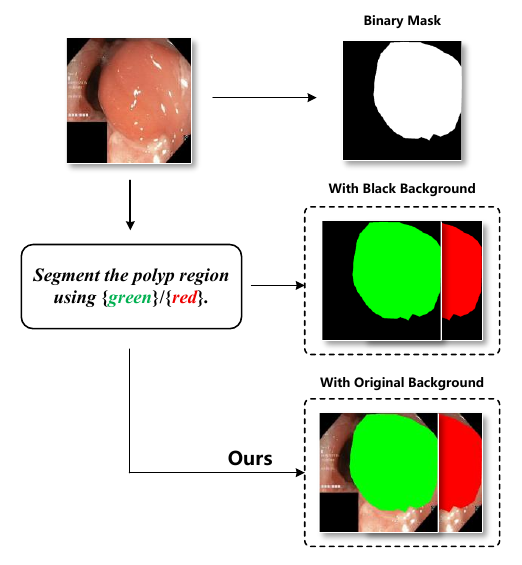}
\caption{Conceptual illustration of the proposed framework. Traditional methods generate binary masks. Our method reformulates segmentation as an image editing task. It renders color-coded regions on the original image to align with generative priors and mitigate domain shift.}
\label{fig:comparison1}
\end{figure}

We propose InstEditSeg, a unified generative framework that reformulates segmentation as instruction-driven image editing, producing an edited image that keeps the original content and adds color-coded regions; this aligns the output with the natural image prior of latent diffusion models and reduces the mismatch with medical images (Fig.~\ref{fig:comparison1}). To compensate for the limited spatial guidance of denoising alone, we introduce DINOv3 as an auxiliary visual encoder and inject hierarchical features from a multi-scale pyramid into the U-Net, providing discriminative priors from local boundaries to global structure.

The editing formulation provides three advantages over discriminative baselines. The color-coded overlay keeps the original image content and aligns with the generative prior, so the model generalizes better to unseen settings, attaining the best Dice on the unseen PolypGen and ISIC2017 sets. The generative formulation also lets the model retain low-contrast secondary lesions that discriminative decoders drop, and a single architecture conditioned only on a text instruction covers multiple categories without task-specific heads or manual prompting. In-domain, the strongest discriminative baselines remain competitive, so the advantage is concentrated in generalization, segmentation completeness, and task control.

The main contributions of this work are summarized as follows:

\begin{itemize}
    \item We propose InstEditSeg, a unified generative framework that recasts medical image segmentation as instruction-driven color-coded image editing, eliminating task-specific heads and aligning the output with latent diffusion priors.

    \item We develop an instruction-based data pipeline that converts segmentation annotations into unified image-text samples via textual instructions and randomized colors.

    \item We design a DINOv3-guided multi-scale feature pyramid that injects hierarchical discriminative priors into the diffusion U-Net, improving lesion localization and boundary delineation.

    \item Experiments on polyp and skin lesion benchmarks show competitive accuracy, notably better cross-domain generalization on unseen data, and concrete advantages of the editing formulation in multi-lesion completeness and instruction-conditioned task control.
\end{itemize}

\section{Related Work}
\label{sec:related_work}

\subsection{Deep Learning for Medical Image Segmentation}
Deep learning has established the dominant paradigms for medical segmentation. U-Net \cite{ronneberger2015u} set the encoder-decoder standard with skip connections, which Attention U-Net \cite{oktay2018attention} and U-Net++ \cite{zhou2018unet} refined with attention and dense connections. Specialized architectures \cite{jha2020resunet, yu2019automated} further targeted polyp and skin lesion segmentation. Transformer-based designs now lead, with Polyp-PVT \cite{dong2023polyp} using pyramid vision transformers and EMCAD \cite{rahman2024emcad} an efficient multi-scale attention decoder. Nevertheless, discriminative models remain sensitive to domain shifts and require many pixel-level annotations, so our work explores a generative paradigm that leverages pretrained diffusion priors for better cross-domain generalization.

\subsection{Foundation Models in Medical Imaging}
Foundation models have reshaped visual understanding. SAM \cite{kirillov2023segany} achieved strong zero-shot segmentation on natural images, prompting medical adaptation such as MedSAM \cite{ma2024segment}, Medical SAM Adapter \cite{wu2023medicalsamadapter}, SAM-Med2D \cite{cheng2023sammed2d}, Polyp-SAM \cite{li2024polyp}, and MedSAM2 \cite{ma2025medsam2} based on SAM 2 \cite{ravi2024sam2}. However, systematic evaluations \cite{huang2024sammedsurvey} show that SAM-family models are unstable on medical data, and their point or box prompts are cumbersome for low-contrast lesions with ambiguous boundaries. In contrast, we generate results directly from a textual instruction without manual prompting.

\subsection{Diffusion Models for Segmentation}
Diffusion probabilistic models \cite{ho2020denoising} are now widely applied to segmentation. SegDiff \cite{amit2021segdiff} introduced iterative denoising for segmentation, Wolleb \emph{et al.} \cite{wolleb2022diffusion} obtained uncertainty ensembles, and MedSegDiff \cite{wu2024medsegdiff} and MedSegDiff-V2 \cite{wu2024medsegdiffv2} generalized this across medical tasks, while SDSeg \cite{lin2024stable} and TSLDseg \cite{yang2025tsldseg} operate in the latent space of Stable Diffusion. Beyond mask prediction, segmentation has been cast as generation through Pix2Seq-D \cite{chen2023generalist}, SegGen \cite{ye2024seggen}, LDMSeg \cite{vangansbeke2024ldmseg}, and VPD \cite{zhao2023vpd}, and, in medical imaging, through TextDiff \cite{feng2024textdiff}, DiffDGSS \cite{xie2024diffdgss}, and GenMed \cite{zhang2026genmed}; see \cite{kazerouni2023diffusion} for a survey. Most of these methods still generate standalone binary masks with task-specific objectives, underutilizing the generative priors of latent diffusion models. By contrast, we reformulate segmentation as instruction-driven editing that produces color-coded overlays, aligning with the learned distribution of latent diffusion models \cite{rombach2022high} and reducing the domain gap.

\subsection{Instruction-Driven Editing and Unified Vision Generation}
Instruction-following editing has become a powerful interface for visual generation. InstructPix2Pix \cite{brooks2023instructpix2pix} conditions editing on instructions, and ControlNet \cite{zhang2023adding} and T2I-Adapter \cite{mou2024t2i} inject spatial controls into frozen diffusion models. Unified visual generation models formulate perception as conditional generation, from Pix2Seq \cite{chen2021pix2seq} to diffusion generators such as Vision Banana \cite{visionbanana2026} and SenseNova Vision \cite{sensenova2026sensenovavision}, which solve segmentation, detection, and depth from textual instructions without task-specific heads, and the CLIP-driven universal model \cite{liu2023clipuniversal} shows similar potential in medical imaging. Directly applying these to medical images faces a distribution gap and limited spatial guidance from text alone. Our framework addresses both by injecting a DINOv3-guided multi-scale feature pyramid that complements the instruction-driven formulation.

\section{Methodology}
\label{sec:methodology}
\begin{figure*}
\centering
\includegraphics[width=0.7\linewidth]{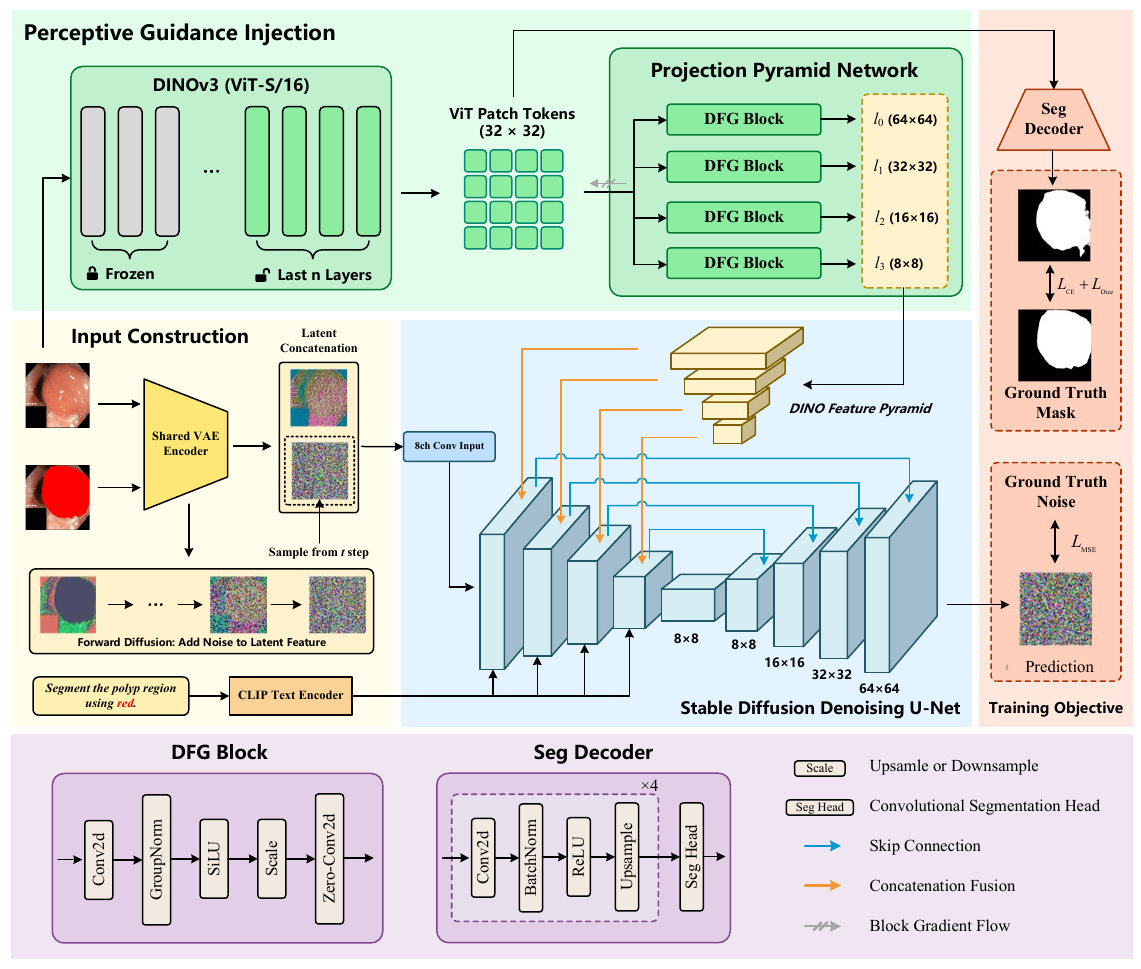}
\caption{Overall architecture of the proposed framework. It integrates a Stable Diffusion U-Net with a DINOv3 guidance branch. The input image is VAE-encoded and concatenated with the noisy latent as the conditioning input. Multi-scale DINOv3 features are projected via DFG blocks and injected into the U-Net. A lightweight auxiliary decoder supervises intermediate features with cross-entropy and Dice losses to ensure anatomical consistency.}
\label{fig:method}
\end{figure*}

The overall architecture of the proposed InstEditSeg framework is illustrated in Fig.~\ref{fig:method}. Unlike conventional segmentation networks that directly predict pixel-wise semantic labels, our method reformulates medical image segmentation as an instruction-driven conditional image editing task. Given an input medical image together with a textual instruction, the diffusion model generates an edited image in which the target region is highlighted using color-coded annotations, thereby achieving segmentation within a unified generative framework.

The framework consists of three components. A latent diffusion model based on Stable Diffusion \cite{rombach2022high} serves as the generative backbone for conditional generation in the latent space. A frozen DINOv3 \cite{simeoni2025dinov3} encoder extracts discriminative visual representations, and a multi-scale feature projection module transforms them into hierarchical visual priors spatially aligned with the decoding stages of the diffusion U-Net. These priors are injected into the denoising process to jointly model global semantics, local anatomy, and fine boundaries, which preserves the generative capability of latent diffusion models while improving segmentation accuracy for medical images.
\subsection{Instruction-Based Segmentation Reformulation}
Given an input image $\mathbf{X}$ and its corresponding ground-truth segmentation mask $\mathbf{M}$, a random color mapping function $\Phi(\cdot)$ is first applied to obtain the color-coded segmentation mask $\mathbf{C}$:
\begin{equation}
\mathbf{C} = \Phi(\mathbf{M}).
\end{equation}
The target edited image $\mathbf{Y}$ is subsequently constructed by overlaying the color-coded mask $\mathbf{C}$ onto the original image $\mathbf{X}$ via a composite function $f(\cdot, \cdot)$:
\begin{equation}
\mathbf{Y} = f(\mathbf{X}, \mathbf{C}).
\end{equation}
Specifically, the overlay renders the target region with the assigned color while preserving the original content outside the target region, so that the underlying anatomical context is retained in the edited image.
Meanwhile, a textual instruction $T$ is randomly sampled from a predefined template set (e.g., \emph{``Segment the polyp region using red.''}). Both the color assignments and the instruction templates are randomly sampled during training. This strategy provides multiple linguistic and visual representations for the same semantic category, thereby enhancing the model's robustness in instruction comprehension.
During training, the diffusion model learns to reconstruct the edited image $\mathbf{Y}$ conditioned on the input image $\mathbf{X}$ and textual instruction $T$. During inference, only the original image along with the specified instruction is required to generate the final color-coded segmentation output.
\subsection{Diffusion-Based Segmentation Framework}
We adopt Stable Diffusion as our unified generative backbone. Given the target edited image $\mathbf{Y}$, a frozen VAE encoder $\mathcal{E}(\cdot)$ first maps it into the latent space:
\begin{equation}
\mathbf{z}_y = \mathcal{E}(\mathbf{Y}).
\end{equation}
Following the standard forward diffusion process, Gaussian noise $\boldsymbol{\epsilon} \sim \mathcal{N}(0, \mathbf{I})$ is added at timestep $t$:
\begin{equation}
\mathbf{z}_t = \sqrt{\bar{\alpha}_t} \mathbf{z}_y + \sqrt{1 - \bar{\alpha}_t} \boldsymbol{\epsilon},
\end{equation}
where $\bar{\alpha}_t$ denotes the cumulative product of the noise schedule coefficients up to step $t$, and $\mathbf{I}$ represents the identity matrix.
Similarly, the input image $\mathbf{X}$ is encoded into its latent representation:
\begin{equation}
\mathbf{z}_I = \mathcal{E}(\mathbf{X}).
\end{equation}
Following the image editing paradigm \cite{brooks2023instructpix2pix}, the latent code $\mathbf{z}_I$ of the input image is concatenated with the noisy latent $\mathbf{z}_t$ along the channel dimension, and this concatenated pair serves as the image conditioning input of the U-Net.
The conditional diffusion U-Net parameterizes the noise prediction function $\boldsymbol{\epsilon}_\theta$. Taking the noisy latent $\mathbf{z}_t$, image condition latent $\mathbf{z}_I$, timestep $t$, text prompt embedding $\mathbf{c}_T$ (encoded from instruction $T$), and DINO-guided visual priors $\mathcal{P}$ as inputs, the model predicts the added noise $\hat{\boldsymbol{\epsilon}}$:
\begin{equation}
\hat{\boldsymbol{\epsilon}} = \boldsymbol{\epsilon}_\theta(\mathbf{z}_t, \mathbf{z}_I, t, \mathbf{c}_T, \mathcal{P}).
\end{equation}
\subsection{DINO-Guided Multi-Scale Feature Projection}
Although diffusion models possess strong generative capabilities, relying solely on standard image and text conditions is often insufficient for accurately capturing subtle anatomical structures and ambiguous lesion boundaries in medical images. To address this limitation, we introduce a frozen DINOv3 encoder as an auxiliary visual backbone to supply discriminative semantic priors throughout the denoising process.
Let $\mathbf{F}_D \in \mathbb{R}^{N \times D}$ denote the sequence of patch token representations extracted by DINOv3, where $N$ is the number of tokens and $D$ is the feature dimension. Because these vision transformer tokens fundamentally differ from the spatial feature maps utilized by the diffusion U-Net, they cannot be directly injected into the denoising network.
To bridge this representation gap, we design a DINO Feature Guidance (DFG) Block. This module progressively transforms transformer tokens into hierarchical convolutional feature maps via token projection, spatial reconstruction, and multi-scale feature transformation, with its detailed architecture illustrated in Fig.~\ref{fig:method}. The resulting visual priors are expressed as:
\begin{equation}
\mathcal{P} = \{\mathbf{P}_1, \mathbf{P}_2, \mathbf{P}_3, \mathbf{P}_4\},
\end{equation}
where the spatial resolutions of $\mathbf{P}_i$ ($i \in \{1, 2, 3, 4\}$) match the four decoding stages of the diffusion U-Net, establishing spatially aligned visual priors for multi-scale feature fusion; the active DINOv3 feature layers are grouped into four levels corresponding to these stages.

\subsection{DINO Feature Injection}

After extracting the multi-scale visual priors $\mathcal{P}$, they are injected into their corresponding stages within the diffusion U-Net.
Existing conditional diffusion models (e.g., ControlNet \cite{zhang2023adding}, T2I-Adapter \cite{mou2024t2i}) incorporate external conditions via element-wise addition, which implicitly assumes that the conditioning and network features share an aligned embedding space. In our framework, diffusion features encode generative representations for noise prediction, whereas DINOv3 provides discriminative semantic representations, so these heterogeneous features exhibit distinct statistical properties.
To address this domain mismatch, we adopt a Concatenation and Zero-Convolution fusion strategy. For the $i$-th feature level, the U-Net feature map $\mathbf{U}_i$ and the corresponding visual prior $\mathbf{P}_i$ are concatenated along the channel dimension:
\begin{equation}
\tilde{\mathbf{U}}_i = \operatorname{Concat}(\mathbf{U}_i, \mathbf{P}_i).
\end{equation}
The concatenated representation is then processed by a zero-initialized convolutional layer $\operatorname{Conv}_{\text{zero}}(\cdot)$:
\begin{equation}
\hat{\mathbf{U}}_i = \operatorname{Conv}_{\text{zero}}(\tilde{\mathbf{U}}_i).
\end{equation}
Compared with additive fusion, concatenation preserves the full information from both branches, enabling the network to learn complex non-linear interactions automatically. The zero initialization of the convolution ensures that the initial forward pass matches the pretrained diffusion model, and because the fused output is added back to the U-Net feature through a residual connection ($\mathbf{U}_i \leftarrow \mathbf{U}_i + \hat{\mathbf{U}}_i$), the pretrained behavior is preserved until the guidance branch takes effect.

\subsection{Training Objective}

The proposed framework is optimized end-to-end using a joint objective. The primary noise prediction loss $\mathcal{L}_{\text{noise}}$ is defined as:
\begin{equation}
\mathcal{L}_{\text{noise}} = \mathbb{E}_{\mathbf{z}_y, \boldsymbol{\epsilon}, t} \left[ \left| \boldsymbol{\epsilon} - \boldsymbol{\epsilon}_\theta(\mathbf{z}_t, \mathbf{z}_I, t, \mathbf{c}_T, \mathcal{P}) \right|_2^2 \right].
\end{equation}
To further enforce anatomical structural consistency, a lightweight auxiliary decoder (shown in Fig.~\ref{fig:method}) is attached to the visual guidance branch. We apply a combination of Dice loss ($\mathcal{L}_{\text{Dice}}$) and Cross-Entropy loss ($\mathcal{L}_{\text{CE}}$) to supervise the intermediate features:
\begin{equation}
\mathcal{L}_{\text{seg}} = \mathcal{L}_{\text{Dice}} + \mathcal{L}_{\text{CE}}.
\end{equation}
The overall optimization objective is given by:
\begin{equation}
\mathcal{L} = \mathcal{L}_{\text{noise}} + \lambda \mathcal{L}_{\text{seg}},
\end{equation}
where $\lambda > 0$ is a balancing hyperparameter that weights the auxiliary segmentation supervision relative to the diffusion denoising loss.

\subsection{Inference Strategy}
\label{subsec:inference}

\begin{figure}
\centering
\includegraphics[width=0.95\linewidth]{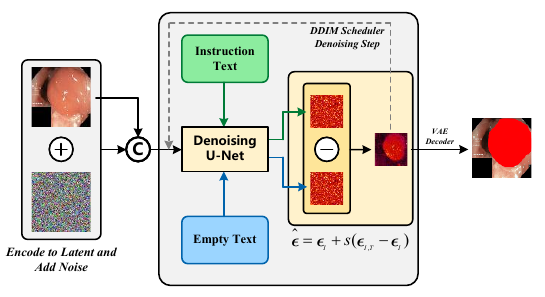}
\caption{Inference sampling pipeline. A dual-branch CFG strategy is adopted to reduce latency. The U-Net takes the instruction text and an empty text to predict the conditional and unconditional noises. The guided noise is computed as Eq.~\eqref{eq:cfg}. DINO feature injection is omitted from the figure for the sake of brevity.}
\label{fig:inference}
\end{figure}

During inference, classifier-free guidance (CFG) is used to control conditional sampling.
Unlike general multi-modal generation models that require three forward passes per step (unconditioned, text-conditioned, and joint image-text conditioned), we propose a simplified dual-branch guidance strategy tailored specifically for medical image segmentation.
Medical image segmentation is driven by visual anatomy, so spatial boundaries are determined by the input image while the text instruction only selects the target category; the image condition alone is therefore a strong structural constraint, and a separate text-only branch adds computation without benefit and can induce semantic drift.
Accordingly, our method requires only two U-Net forward evaluations per denoising step:
Noise prediction under image-only conditioning (with an empty text embedding $\varnothing$):
\begin{equation}
\boldsymbol{\epsilon}_I = \boldsymbol{\epsilon}_\theta(\mathbf{z}_t, \mathbf{z}_I, t, \varnothing, \mathcal{P}).
\end{equation}
Noise prediction under joint image-text conditioning:
\begin{equation}
\boldsymbol{\epsilon}_{I,T} = \boldsymbol{\epsilon}_\theta(\mathbf{z}_t, \mathbf{z}_I, t, \mathbf{c}_T, \mathcal{P}).
\end{equation}
The final guided noise prediction $\hat{\boldsymbol{\epsilon}}$ is formulated as:
\begin{equation}
\hat{\boldsymbol{\epsilon}} = \boldsymbol{\epsilon}_I + s (\boldsymbol{\epsilon}_{I,T} - \boldsymbol{\epsilon}_I),
\label{eq:cfg}
\end{equation}
where $s$ denotes the guidance scale factor for text control.
By restricting text guidance strictly to task specification while relying on image features for spatial alignment, this dual-branch scheme preserves high image fidelity and reduces inference latency compared to standard three-branch CFG sampling.

\section{Experiments}
\label{sec:experiments}

\subsection{Experimental Setup}

\subsubsection{Datasets}
To comprehensively validate the effectiveness of the proposed method, we conducted experiments on two medical image segmentation tasks: polyp segmentation and skin lesion segmentation. Polyp segmentation evaluates the model's capability for target localization and boundary delineation within complex endoscopic scenes, while skin lesion segmentation verifies the generalization ability of the unified generative framework across a different medical imaging modality.

\begin{table}[ht]
\centering
\caption{Statistics of the datasets used in the experiments. ``Unseen Test'' indicates that the dataset is not involved in any training stage.}
\label{tab:datasets}
\renewcommand{\arraystretch}{1.2}
\resizebox{0.8\columnwidth}{!}{%
\begin{tabular}{@{}lcccc@{}}
\toprule
Dataset & Task & Train & Test & Unseen Test \\
\midrule
Kvasir-SEG          & Polyp & 900  & 100 & $\times$ \\
CVC-ClinicDB        & Polyp & 550  & 62 & $\times$ \\
CVC-ColonDB         & Polyp & 341  & 38  & $\times$ \\
ETIS-LaribPolypDB   & Polyp & 176  & 20  & $\times$ \\
PolypGen            & Polyp & --   & 1411 & $\checkmark$ \\
\midrule
ISIC2016            & Skin  & 900  & 379 & $\times$ \\
ISIC2017            & Skin  & --   & 600 & $\checkmark$ \\
\bottomrule
\end{tabular}%
}
\end{table}

\begin{table*}[ht]
\centering
\caption{Quantitative comparison of polyp segmentation performance. Bold denotes the best result among all compared methods.}
\label{tab:polyp}
\resizebox{0.8\textwidth}{!}{%
\begin{tabular}{@{}l|c|cc|cc|cc|cc|cc@{}}
\toprule
\multirow{2}{*}{Method} & \multirow{2}{*}{Prompt} & \multicolumn{2}{c|}{Kvasir-SEG} & \multicolumn{2}{c|}{ETIS} & \multicolumn{2}{c|}{CVC-ClinicDB} & \multicolumn{2}{c|}{CVC-ColonDB} & \multicolumn{2}{c}{PolypGen (\textit{Unseen})} \\ 

& & Dice & IoU & Dice & IoU & Dice & IoU & Dice & IoU & Dice & IoU \\ \midrule
U-Net             & None & 81.99 & 73.81 & 69.64 & 61.84 & 82.03 & 74.92 & 78.76 & 63.04 & 69.57 & 56.21 \\
Polyp-PVT         & None & 91.86 & 86.52 & 88.70 & 79.33 & 93.36 & 88.59 & 88.31 & 71.25 & 80.84 & 75.88 \\
EMCAD             & None & \textbf{93.74} & \textbf{89.21} & 88.46 & 82.62 & \textbf{93.22} & \textbf{88.04} & 89.48 & 83.63 & 80.78 & 74.25 \\
\midrule
SAM               & bbox & 89.25 & 82.54 & 77.01 & 68.88 & 89.72 & 83.57 & 75.32 & 66.69 & 80.94 & 72.20 \\
MedSAM            & bbox & 92.75 & 88.14 & \textbf{92.61} & \textbf{87.28} & 92.31 & 86.24 & 87.78 & 79.01 & 83.51 & 74.59 \\
\midrule
MedSegDiff        & None & 87.98 & 80.19 & 84.81 & 75.38 & 87.26 & 79.95 & 87.05 & 79.78 & 75.25 & 66.17 \\
MedSegDiff-V2     & None & 92.12 & 87.43 & 89.36 & 81.35 & 92.02 & 86.79 & 91.22 & 84.41 & 80.74 & 75.82 \\
SDSeg             & None & 90.67 & 87.56 & 90.86 & 84.51 & 89.73 & 81.57 & 90.17 & 85.44 & 82.46 & 76.03 \\
TSLDseg           & None & 91.98 & 87.01 & 91.44 & 85.27 & 90.25 & 85.09 & 91.82 & 86.76 & 79.56 & 71.64 \\
\midrule
\textbf{InstEditSeg} & Text & 92.10 & 87.05 & 92.01 & 85.93 & 92.83 & 87.32 & \textbf{92.95} & \textbf{87.46} & \textbf{83.92} & \textbf{77.50} \\
\bottomrule
\end{tabular}%
}
\end{table*}

\textbf{Polyp Segmentation.} We utilized four public datasets: Kvasir-SEG~\cite{jha2019kvasir}, CVC-ClinicDB~\cite{bernal2015wm}, CVC-ColonDB~\cite{tajbakhsh2015automated}, and ETIS-LaribPolypDB~\cite{silva2014toward}; Table~\ref{tab:datasets} summarizes their sample sizes. Following common practice, the training sets of these four datasets are combined, with independent evaluations on each test set. To rigorously assess cross-dataset generalization, PolypGen~\cite{ali2023multi} is employed as an unseen test set: it was excluded from training, and its images originate from different medical institutions and endoscopic equipment, presenting significant domain shifts that provide an objective benchmark for generalization.

\textbf{Skin Lesion Segmentation.} We performed training on ISIC2016 and testing on both ISIC2016~\cite{gutman2016skin} and ISIC2017~\cite{codella2018skin}. As shown in Table~\ref{tab:datasets}, ISIC2017 was held out during training and served solely to evaluate cross-dataset generalization.

For both tasks, training samples were constructed with the instruction-based pipeline proposed in Section~\ref{sec:methodology}: text templates and color codings were randomly sampled to build unified image-text pairs. No additional annotation data or external training samples were introduced. All compared methods utilize their public implementations and official recommended configurations. To ensure a strictly controlled comparison, the discriminative baseline EMCAD was retrained from scratch on the same training data (the combined polyp training sets and the ISIC2016 training set, respectively) following its official implementation, including multi-scale test-time augmentation, and all of its reported numbers are from this controlled re-run.

\subsubsection{Implementation Details}
All experiments were implemented in PyTorch and trained on a single NVIDIA A100 (40~GB) GPU.

We adopted Stable Diffusion as the generative backbone. All images were resized to $512 \times 512$. The VAE encoder and the DINOv3 backbone remained frozen throughout training, while only the diffusion U-Net and the CLIP Text Encoder were fine-tuned. The initial learning rate was set to $1\times10^{-4}$ for the U-Net and $5\times10^{-7}$ for the CLIP Text Encoder. We employed the AdamW optimizer with a batch size of 8 and a weight decay of $1\times10^{-3}$. The total training steps were set to 20,000. The DINOv3 ViT-S/16 model was adopted as the default visual encoder, with 6 active feature layers used to construct the multi-scale guidance pyramid, and the auxiliary loss weight $\lambda$ was set to 0.3.

The model was supervised by the constructed image editing target, jointly optimizing the diffusion noise prediction and auxiliary segmentation losses. Unless otherwise specified, all experiments adopted identical training configurations.

During inference, we utilized the DDIM scheduler \cite{song2021denoising} with 25 sampling steps and the proposed dual-branch classifier-free guidance (CFG) strategy described in Section~\ref{subsec:inference}, with the guidance scale $s$ set to 7.5. The Dice Similarity Coefficient (Dice) and Intersection over Union (IoU) were adopted as the primary evaluation metrics.

\subsection{Comparison with State-of-the-art Methods}

We compared the proposed method against several state-of-the-art segmentation approaches, including traditional CNN-based methods (U-Net~\cite{ronneberger2015u}), transformer-based methods (Polyp-PVT~\cite{dong2023polyp}, EMCAD~\cite{rahman2024emcad}), foundation models (SAM~\cite{kirillov2023segany}, MedSAM~\cite{ma2024segment}), and diffusion-based methods (MedSegDiff~\cite{wu2024medsegdiff}, MedSegDiff-V2~\cite{wu2024medsegdiffv2}, SDSeg~\cite{lin2024stable}, TSLDseg~\cite{yang2025tsldseg}).
Table~\ref{tab:polyp} reports the polyp segmentation results. Among the baselines, the discriminative EMCAD attains the highest Dice and IoU on Kvasir-SEG and CVC-ClinicDB (93.74\% and 93.22\%, respectively); the foundation model MedSAM is strongest on ETIS, with a Dice of 92.61\%; and the best diffusion baseline, TSLDseg, reaches a Dice of 91.82\% on CVC-ColonDB.

Our proposed method achieves the best overall results on CVC-ColonDB, with a Dice of 92.95\% and an IoU of 87.46\%. On ETIS it attains a Dice of 92.01\%, ranking second only behind MedSAM, and on Kvasir-SEG and CVC-ClinicDB it delivers highly competitive results, with 92.10\% and 92.83\% Dice. Most importantly, on the unseen PolypGen dataset it achieves the best Dice of 83.92\% and IoU of 77.50\%, outperforming all compared methods and demonstrating strong cross-domain generalization. Paired Wilcoxon tests against the strongest baseline confirm that this advantage is significant on CVC-ColonDB with $p<0.05$ and on PolypGen with $p<0.001$, that it is statistically tied ($p>0.05$) on ETIS and CVC-ClinicDB, and that EMCAD remains ahead on Kvasir-SEG with $p<0.05$.

\begin{table}[t]
\centering
\caption{Quantitative comparison of skin lesion segmentation performance. Bold denotes the best result among all compared methods.}
\label{tab:skin}
\renewcommand{\arraystretch}{1.2}
\begin{tabular}{@{}l|cc|cc@{}}
\toprule
\multirow{2}{*}{Method} & \multicolumn{2}{c|}{ISIC2016} & \multicolumn{2}{c}{ISIC2017 (\textit{Unseen})} \\ 
& Dice & IoU & Dice & IoU \\ \midrule
U-Net               & 85.45 & 75.52 & 77.35 & 68.97 \\
Polyp-PVT           & 88.75 & 81.10 & 81.76 & 72.58 \\
EMCAD               & 92.65 & 87.40 & 82.39 & 74.88 \\
\midrule
SAM                 & 88.50 & 80.86 & 78.08 & 69.12 \\
MedSAM              & 90.24 & 82.74 & 82.83 & \textbf{76.55} \\
\midrule
MedSegDiff          & 88.13 & 75.17 & 72.34 & 67.24 \\
MedSegDiff-V2       & 92.35 & 87.06 & 80.87 & 74.41 \\
SDSeg               & \textbf{92.73} & \textbf{88.68} & 76.32 & 68.03 \\
TSLDseg             & 92.08 & 88.22 & 80.51 & 71.69 \\
\midrule
\textbf{InstEditSeg} & 92.58 & 86.91 & \textbf{83.14} & 75.62 \\
\bottomrule
\end{tabular}
\end{table}

\begin{figure*}[t]
\centering
\includegraphics[width=0.4\linewidth]{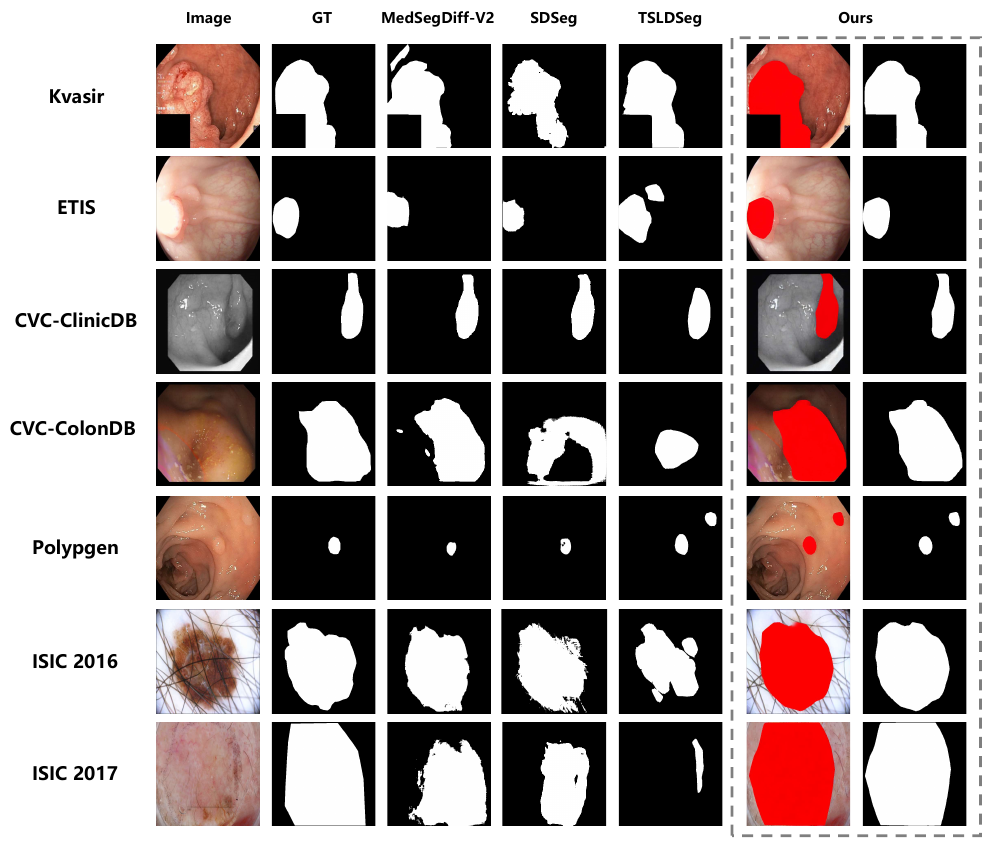}
\caption{Qualitative comparison of representative methods on challenging polyp and skin lesion cases. Our method delineates lesion boundaries more precisely in low-contrast, irregular, or ambiguous cases, where other methods often produce fragmented predictions.}
\label{fig:visual_results}
\end{figure*}

Table~\ref{tab:skin} presents the skin lesion segmentation results. On the in-domain ISIC2016 test set, the diffusion-based SDSeg leads with a Dice of 92.73\% and an IoU of 88.68\%, closely followed by the discriminative EMCAD (92.65\%) and our method (92.58\%), both above the remaining diffusion baselines and the foundation model MedSAM. This confirms that the instruction-driven editing paradigm captures lesion boundaries in dermoscopic images while remaining competitive with the best discriminative baseline.

On the unseen ISIC2017 dataset, all methods degrade under domain shift. Our framework attains the best Dice of 83.14\% and a competitive IoU of 75.62\%, outperforming the strongest discriminative baseline EMCAD at 82.39\% and the discriminative transformer Polyp-PVT at 81.76\%, with a significant paired Wilcoxon lead over EMCAD at $p<0.05$, while MedSAM holds the highest IoU of 76.55\%. Our framework also degrades less than EMCAD from the in-domain result. These results show that the instruction-driven editing paradigm is robust to domain shift on an unseen skin set, even though the strongest discriminative baselines remain competitive.

Fig.~\ref{fig:visual_results} provides qualitative comparisons on challenging cases from both datasets. For polyp segmentation, our method delineates lesion boundaries with higher precision in cases with low contrast, irregular shapes, or small flat polyps, where other approaches often produce fragmented predictions or miss subtle regions. In skin lesion segmentation, the proposed framework accurately captures lesions with fuzzy borders, while baseline diffusion models tend to over-segment or include background artifacts. These visual results corroborate the quantitative improvements in Tables~\ref{tab:polyp} and \ref{tab:skin}, highlighting the benefit of DINOv3 discriminative priors and the instruction-driven editing formulation.

\subsection{Advantages over Discriminative Segmentation}
\label{sec:generative_advantages}

To answer why a generative editing formulation is preferable to conventional discriminative segmentation, we compare the proposed method against the strongest discriminative baseline, EMCAD, on the axes where the two paradigms materially differ. We focus on cross-domain generalization, segmentation completeness on multi-lesion scenes, instruction-conditioned task control, and sampling flexibility, and we report both the strengths and the regimes in which discriminative methods remain competitive.

\subsubsection{Cross-Domain Generalization}
The clearest advantage of the editing formulation is robustness to domain shift. On the unseen PolypGen set, the framework attains the best Dice and IoU among all compared methods, 83.92\% and 77.50\%, exceeding both the discriminative EMCAD and the medical foundation model MedSAM, and on the unseen ISIC2017 set it attains the best Dice of 83.14\%, outperforming EMCAD and Polyp-PVT and led in IoU only by MedSAM. This generalization advantage is consistent with the design-level ablations, where the overlay representation and the DINOv3 guidance branch provide the largest gains, and it is most pronounced on the unseen PolypGen and ISIC2017 sets. We attribute the behavior to two properties of the editing formulation. First, rendering a color-coded overlay on the original image keeps the generated output close to the natural image distribution, so the pretrained prior remains effective on medical inputs. Second, the DINOv3 feature injection anchors the generator to genuine anatomical boundaries, which reduces the sensitivity of the output to a specific imaging protocol.

\subsubsection{Multi-Lesion Completeness}
\begin{figure}
\centering
\includegraphics[width=\linewidth]{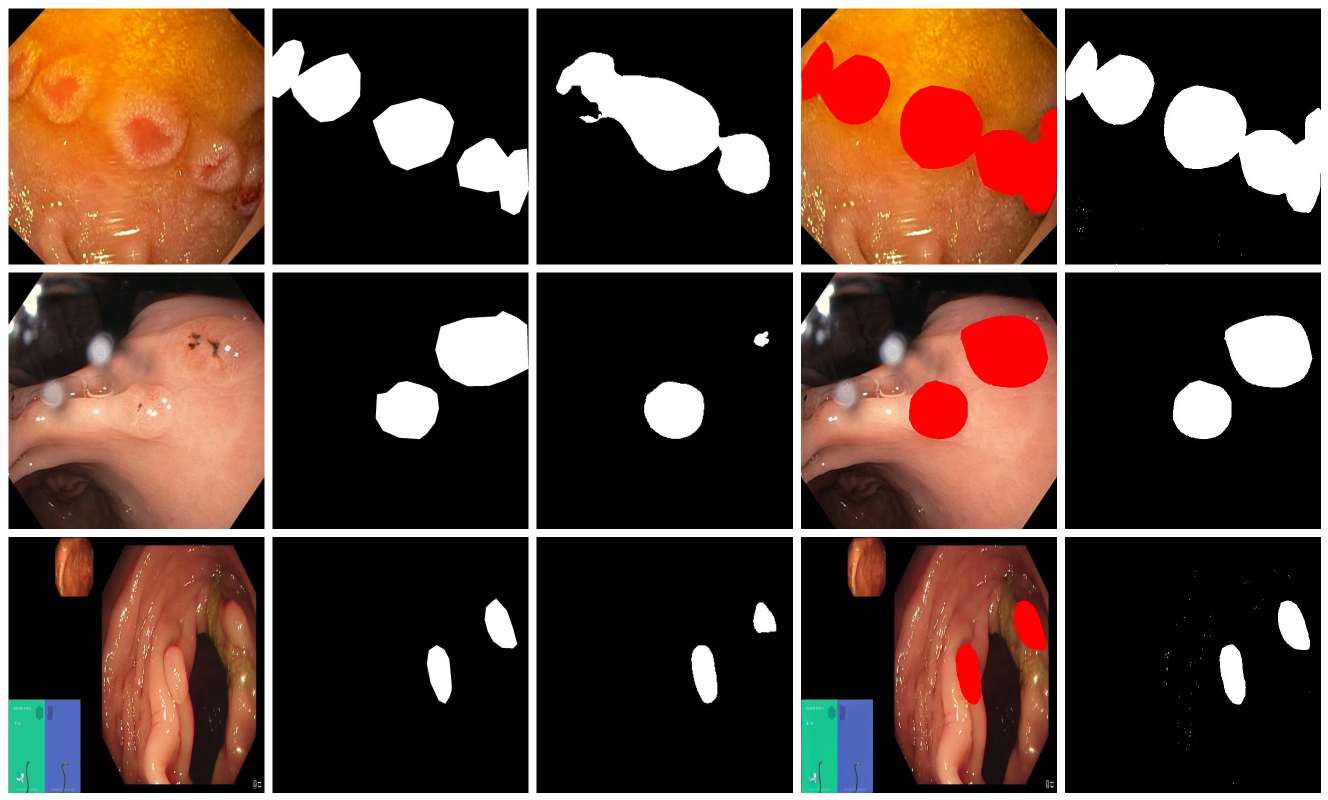}
\caption{Multi-lesion segmentation on PolypGen. Columns: input image, ground truth, InstEditSeg prediction, EMCAD prediction. EMCAD tends to drop secondary lesions, whereas the proposed method recovers multiple lesions.}
\label{fig:multilesion}
\end{figure}

Polyps often appear in multiples, a scenario where small discriminative decoders tend to miss secondary lesions. To quantify this, we select the 119 PolypGen test images whose ground truth contains at least two lesion components (second-largest component $\geq 200$~px) and evaluate both models on all lesions. As reported in Table~\ref{tab:multilesion}, InstEditSeg attains a Dice of 75.93\%, outperforming EMCAD by 12.75 Dice points; qualitative examples are shown in Fig.~\ref{fig:multilesion}. We attribute this primarily to the position-agnostic formulation of our editing task. Discriminative decoders learn an implicit prior about where a lesion is likely to appear, biasing them toward the dominant lesion and suppressing secondary or atypically placed ones. Our framework never anchors the model to a location: the editing target is defined only by the color-coded region and the instruction, and the generator is supervised over the whole image without any position term. It therefore reasons over all candidate regions and preserves multiple lesions regardless of position. DINOv3 guidance helps with low-contrast boundaries, but the absence of an explicit location prior is the principal reason secondary lesions are retained. This position-agnostic behavior is two-sided, since the model also does not parse spatial or size modifiers, such as \emph{``segment only the largest polyp''}, or \emph{``segment the rightmost lesion''}.

\begin{table}[t]
\centering
\caption{Segmentation performance on the 119 PolypGen test images containing multiple polyps (all lesions evaluated).}
\label{tab:multilesion}
\renewcommand{\arraystretch}{1.2}
\begin{tabular}{@{}l|cc@{}}
\toprule
Method & Dice & IoU \\ \midrule
EMCAD               & 63.18 & 53.00 \\
\textbf{InstEditSeg} & \textbf{75.93} & \textbf{64.93} \\ \bottomrule
\end{tabular}
\end{table}

\subsubsection{Unified Instruction-Conditioned Task Control}
A discriminative network is specialized to a single task and output space, whereas InstEditSeg is a single generative architecture that produces a segmentation whenever a textual instruction is provided. The same architecture, without any task-specific head, handles both polyp and skin lesion segmentation, with the instruction selecting the target category; each task is trained separately on its own data. This differs from promptable foundation models, which need a point or box prompt, and from discriminative networks, which require a dedicated decoder per task. The current model does not yet parse spatial or size modifiers, and we treat attribute-conditioned selection as open future work.

\subsubsection{Sampling Flexibility and CFG Analysis}
\begin{figure}
\centering
\includegraphics[width=0.5\linewidth]{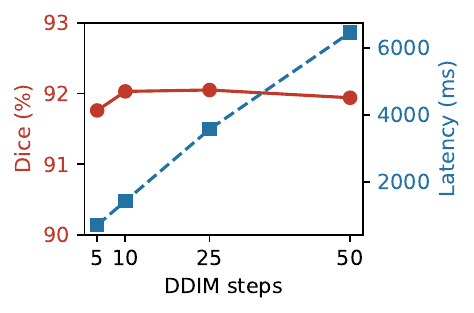}
\caption{Dice accuracy and per-image latency as functions of the number of DDIM sampling steps on Kvasir-SEG.}
\label{fig:efficiency}
\end{figure}

Diffusion sampling is typically the main computational cost of generative segmentation. We analyze the trade-off between DDIM steps and accuracy on the Kvasir-SEG test set. As shown in Table~\ref{tab:efficiency} and Fig.~\ref{fig:efficiency}, five DDIM steps attain a Dice of 91.78\% at 711~ms per image. For comparison, the discriminative baseline EMCAD reaches a Dice of 93.74\% at only 218~ms per image, so the generative pipeline is inherently more costly at inference. Furthermore, we compare inference variants at 25 steps: the proposed dual-branch CFG attains 92.10\% Dice at 3.58~s, a no-CFG variant that performs a single image-and-text conditioned forward per step attains 89.46\% at 1.84~s, and an image-only variant with empty text collapses to 72.75\% at 1.80~s. The no-CFG variant therefore cuts latency by roughly half at a cost of about 2.6 Dice points relative to dual-branch CFG, whereas the image-only variant collapses, which confirms that textual instructions are indispensable for task specification. These results substantiate the efficiency claim of the dual-branch scheme and indicate that, in latency-critical deployments, the step count can be reduced to five with a small accuracy loss, although dropping CFG altogether incurs a more noticeable drop. Increasing the step count to 50 yields 91.94\%, slightly below the 25-step result, so 25 steps was adopted as the default.

\begin{table}[t]
\centering
\caption{Inference efficiency on Kvasir-SEG (100 test images, A100). Latency includes the full generation pipeline per image.}
\label{tab:efficiency}
\renewcommand{\arraystretch}{1.2}
\begin{tabular}{@{}lcc@{}}
\toprule
Configuration & Dice (\%) & Latency (ms) \\ \midrule
EMCAD & \textbf{93.74} & \textbf{218} \\
DDIM 5 steps (dual CFG)  & 91.78 & 711 \\
DDIM 10 steps (dual CFG) & 92.02 & 1428 \\
DDIM 25 steps (dual CFG) & 92.10 & 3575 \\
DDIM 50 steps (dual CFG) & 91.94 & 6461 \\
\midrule
25 steps, no CFG (text$+$image) & 89.46 & 1842 \\
25 steps, image only (empty text) & 72.75 & 1801 \\ \bottomrule
\end{tabular}
\end{table}

\subsection{Ablation Studies}

To verify the contribution of each component, we conducted ablation studies on the Kvasir-SEG and ISIC2016 datasets. The results are summarized in Table~\ref{tab:ablation}.

\begin{table}[t]
\centering
\caption{Ablation study on key components. ``DFG'' denotes the DINO Feature Guidance Block. ``Aux. Loss'' represents the auxiliary segmentation loss.}
\label{tab:ablation}
\renewcommand{\arraystretch}{1.2}
\resizebox{0.9\columnwidth}{!}{%
\begin{tabular}{@{}ccc|cc|cc@{}}
\toprule
\multirow{2}{*}{Baseline} & \multirow{2}{*}{DFG} & \multirow{2}{*}{Aux. Loss} & \multicolumn{2}{c|}{Kvasir-SEG} & \multicolumn{2}{c}{ISIC2016} \\
& & & Dice & IoU & Dice & IoU \\ \midrule
\checkmark & & & 85.47 & 78.82 & 87.64 & 80.08 \\
\checkmark & \checkmark & & 87.73 & 79.59 & 89.12 & 82.33 \\
\checkmark & \checkmark & \checkmark & \textbf{92.10} & \textbf{87.05} & \textbf{92.58} & \textbf{86.91} \\
\bottomrule
\end{tabular}%
}
\end{table}

\textbf{Effectiveness of Components.} The baseline Stable Diffusion model fine-tuned with the editing target attains 85.47\% Dice on Kvasir-SEG. Injecting the DINO Feature Guidance (DFG) block alone raises it to 87.73\%, validating that multi-scale discriminative priors are essential for recovering fine lesion structures. Further incorporating the auxiliary segmentation loss yields the full model performance of 92.10\% Dice, a substantial improvement of +4.37\% over the DFG-only variant. This progressive enhancement demonstrates that both discriminative visual priors and explicit structural supervision are critical in a diffusion framework.

\textbf{Output Representation.} A core design choice of InstEditSeg is predicting a color-coded overlay rendered on the original image rather than a standalone binary mask. We ablate this choice by training two variants under an otherwise identical protocol: (i) a grayscale binary mask output ($[0,1]$), whose instruction removes all color specification, and (ii) a color-coded mask on a pure black background, which keeps the color instruction but discards the image context. The quantitative comparison is reported in Table~\ref{tab:output_repr}.

\begin{table}[t]
\centering
\caption{Ablation study on the output representation. ``Overlay'' is the proposed color-coded mask rendered on the original image. ``Black-bg mask'' is the color-coded mask on a black background. ``Binary mask'' is the grayscale $[0,1]$ mask without color instructions. For the black-background variant, the evaluation color map is remapped to the dominant color actually painted by the model, so its reported scores are corrected Dice.}
\label{tab:output_repr}
\renewcommand{\arraystretch}{1.2}
\resizebox{0.9\columnwidth}{!}{%
\begin{tabular}{@{}l|cc|cc@{}}
\toprule
\multirow{2}{*}{Output Representation} & \multicolumn{2}{c|}{Kvasir-SEG} & \multicolumn{2}{c}{ISIC2016} \\
& Dice & IoU & Dice & IoU \\ \midrule
Binary mask   & 82.35 & 74.22 & 52.60 & 43.22 \\
Black-bg color mask   & 65.31 & 56.99 & 53.40 & 44.88 \\
Overlay (Ours)        & \textbf{92.10} & \textbf{87.05} & \textbf{92.58} & \textbf{86.91} \\ \bottomrule
\end{tabular}%
}
\end{table}

\begin{figure}
\centering
\includegraphics[width=\columnwidth]{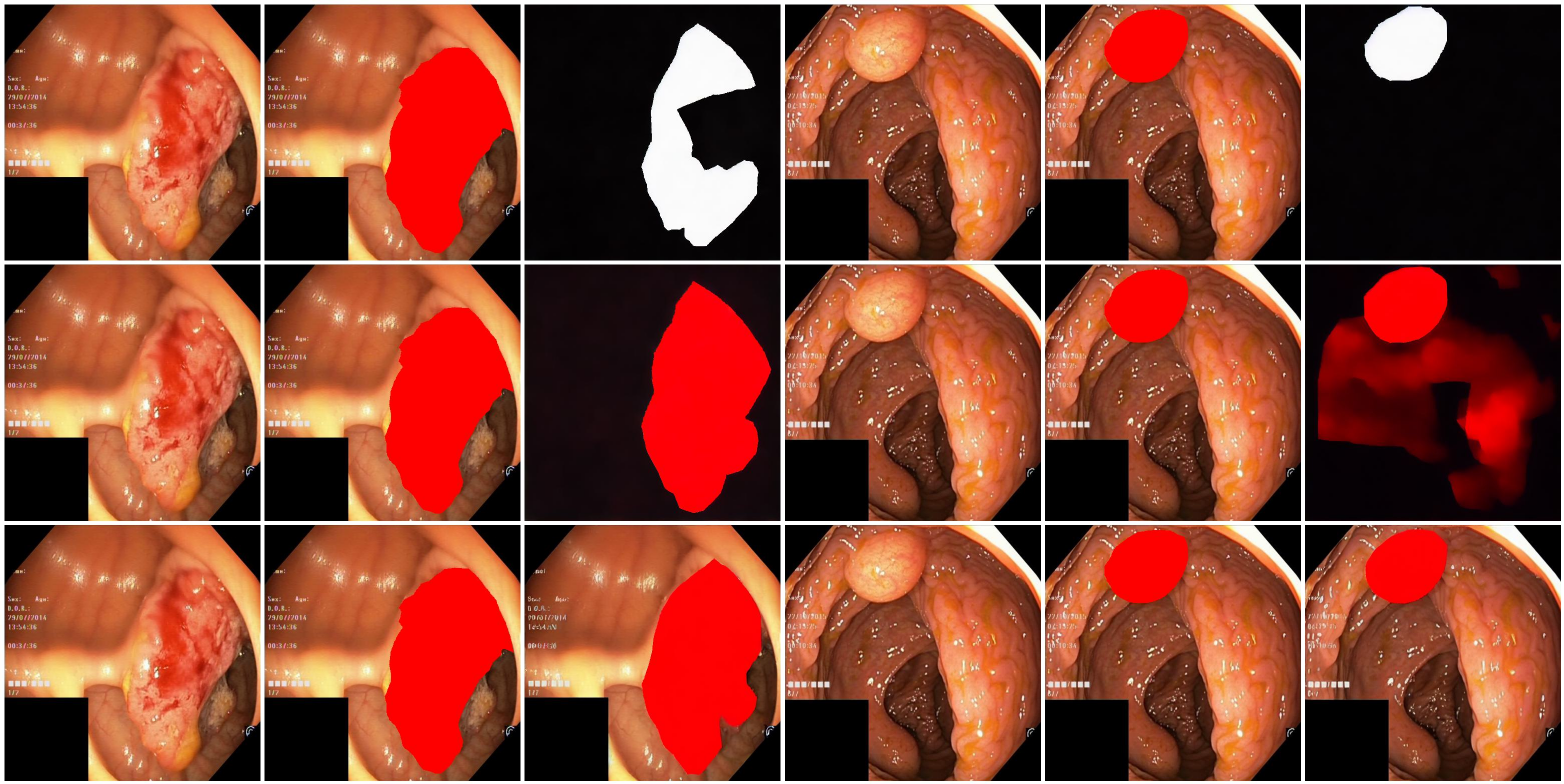}
\caption{Output-representation comparison on two Kvasir-SEG samples. Rows: binary $[0,1]$ mask, color mask on black background, and the proposed overlay. Columns: input image, editing target, model prediction.}
\label{fig:output_repr}
\end{figure}

As shown in Table~\ref{tab:output_repr}, the overlay representation consistently outperforms both alternatives. The binary-mask variant attains only 82.35\% Dice on Kvasir-SEG and 52.60\% on ISIC2016, with a paired Wilcoxon test giving $p<0.001$ against the overlay. This indicates that direct $[0,1]$ mask prediction discards the color-coded structure that latent diffusion models are naturally biased to generate, and that color words in the instruction serve as an additional task-anchoring signal. The black-background variant also degrades substantially even after isolating color-mapping artifacts, reaching 65.31\% corrected Dice on Kvasir-SEG and 53.40\% on ISIC2016, which shows that rendering the mask over the original content provides essential visual context for the denoising process, as illustrated in Fig.~\ref{fig:output_repr}. 

\begin{figure}
\centering
\includegraphics[width=0.85\linewidth]{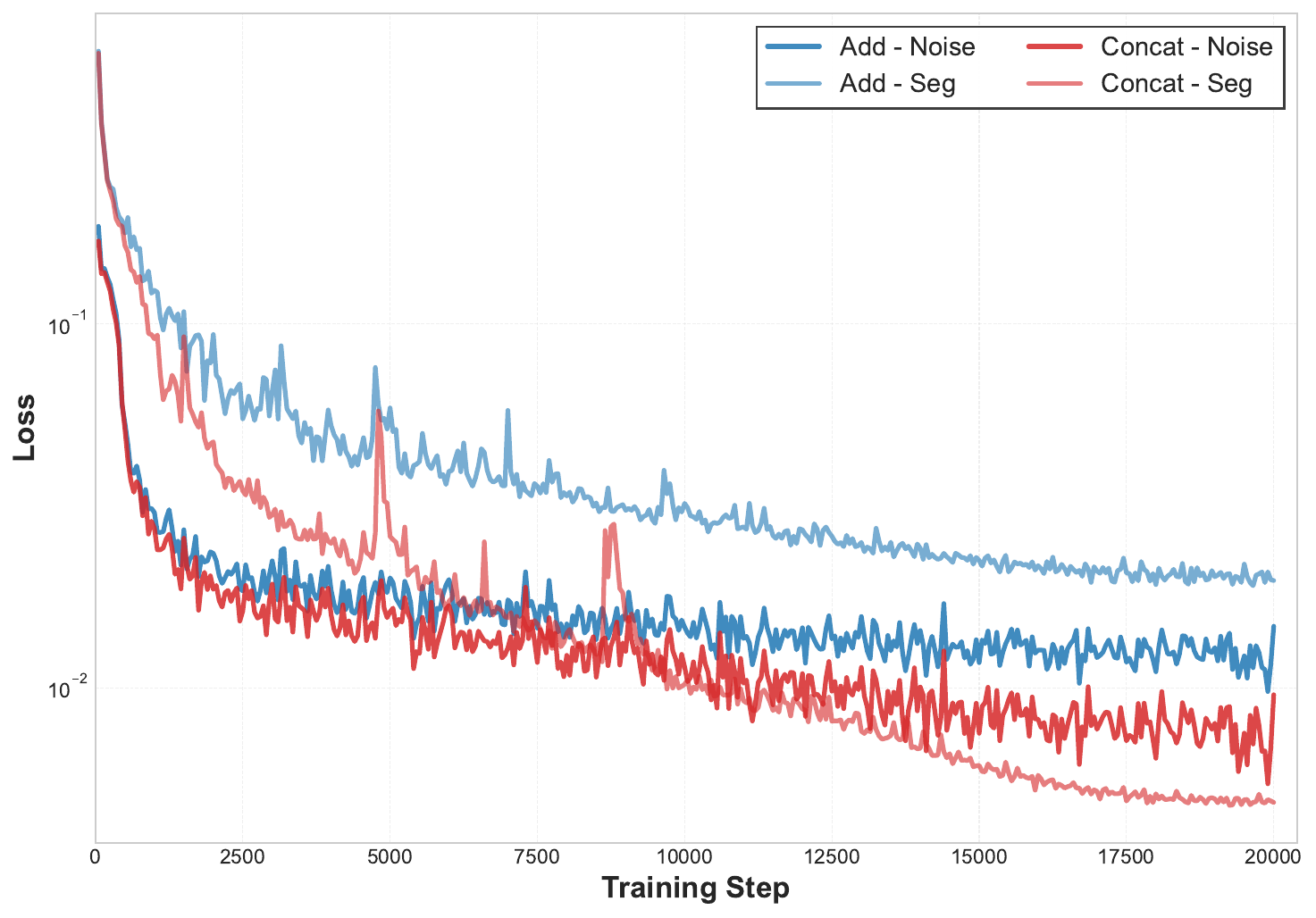}
\caption{Training loss curves of the two fusion strategies.}
\label{fig:loss_curves}
\end{figure}

\textbf{Fusion Strategy.} We compare the proposed concatenation-based fusion against the commonly used element-wise addition. As reported in Table~\ref{tab:ablation_fusion}, with the DFG guidance branch and the auxiliary loss kept identical, concatenation outperforms addition by 0.79\% in Dice on Kvasir-SEG and by 0.81\% on ISIC2016, suggesting that preserving the full feature information from heterogeneous sources allows more effective non-linear interactions, whereas additive fusion forces an overly restrictive alignment between generative and discriminative feature spaces. As shown in Fig.~\ref{fig:loss_curves}, the training loss curve of concatenation fusion remains consistently lower throughout optimization, demonstrating more effective convergence.

\begin{table}[t]
\centering
\caption{Ablation study on feature fusion strategy.}
\label{tab:ablation_fusion}
\renewcommand{\arraystretch}{1.2}
\resizebox{0.8\columnwidth}{!}{%
\begin{tabular}{@{}l|cc|cc@{}}
\toprule
\multirow{2}{*}{Fusion Method} & \multicolumn{2}{c|}{Kvasir-SEG} & \multicolumn{2}{c}{ISIC2016} \\
& Dice & IoU & Dice & IoU \\ \midrule
Element-wise Addition & 91.31 & 85.26 & 91.77 & 84.65 \\
Concatenation         & \textbf{92.10} & \textbf{87.05} & \textbf{92.58} & \textbf{86.91} \\ \bottomrule
\end{tabular}%
}
\end{table}

\begin{figure}
\centering
\includegraphics[width=0.85\linewidth]{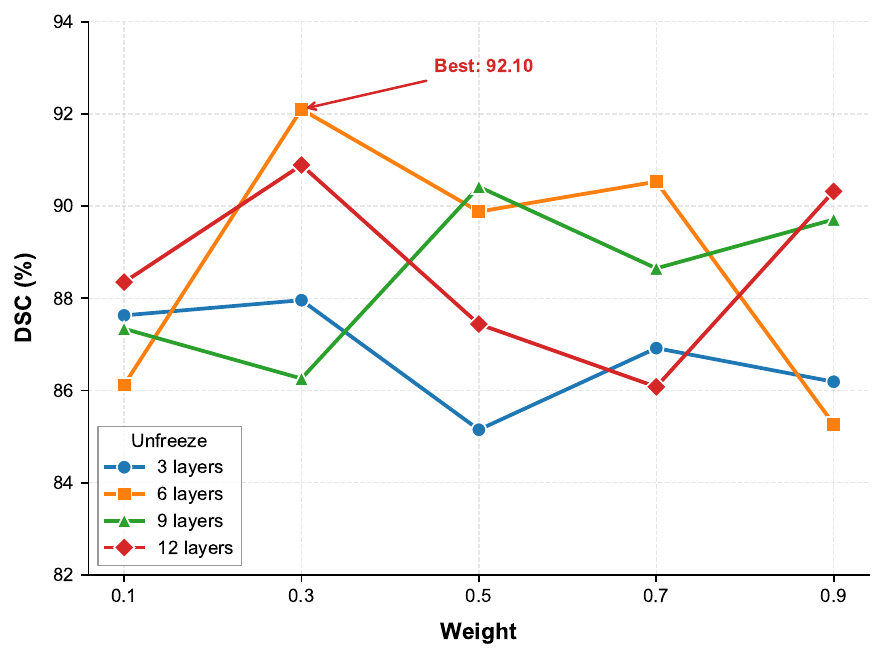}
\caption{Sensitivity analysis of the auxiliary loss weight $\lambda$ and the number of active DINOv3 feature layers.}
\label{fig:hyperparam}
\end{figure}

\textbf{Alternative Auxiliary Backbones.} To verify that the advantage of the DFG branch stems from the DINOv3 representation itself rather than the presence of an auxiliary branch per se, we replace the DINOv3 encoder with alternative visual backbones under an otherwise identical training protocol (same pyramid adapters and auxiliary segmentation head; the last 6 blocks of each alternative backbone are unfrozen, except for ResNet-50, which is fully frozen): a fully supervised ImageNet ResNet-50, a MedSAM image encoder (ViT-B), a CLIP ViT-B/16 image encoder, and a self-supervised DINOv2 ViT-B/14. A variant without any auxiliary branch serves as the lower bound. The comparison is reported in Table~\ref{tab:backbone_replacement}.

\begin{table}[t]
\centering
\caption{Ablation study on the auxiliary backbone of the DFG branch. All backbones share the same pyramid adapter and auxiliary segmentation head. The last 6 blocks of each alternative backbone are unfrozen during training (ResNet-50 is fully frozen), while DINOv3 remains frozen as in the main configuration.}
\label{tab:backbone_replacement}
\renewcommand{\arraystretch}{1.2}
\resizebox{0.9\columnwidth}{!}{%
\begin{tabular}{@{}l|cc|cc@{}}
\toprule
\multirow{2}{*}{Backbone} & \multicolumn{2}{c|}{Kvasir-SEG} & \multicolumn{2}{c}{ISIC2016} \\
& Dice & IoU & Dice & IoU \\ 
\midrule
None & 86.23 & 79.36 & 91.62 & 85.44 \\
ResNet-50 & 85.59 & 78.58 & 91.04 & 85.08 \\
MedSAM ViT-B & 90.60 & 84.65 & 92.22 & 86.35 \\
CLIP ViT-B/16 & 87.51 & 81.08 & 91.77 & 85.86 \\
DINOv2 ViT-B/14 & 91.00 & 85.49 & 91.48 & 85.46 \\
\midrule
DINOv3 ViT-S/16 & 92.10 & 87.05 & 92.58 & 86.91 \\
DINOv3 ViT-B/16 & 92.36 & 87.87 & 92.74 & 87.30 \\
DINOv3 ViT-L/16 & \textbf{92.54} & \textbf{88.11} & \textbf{93.06} & \textbf{87.62} \\ 
\bottomrule
\end{tabular}%
}
\end{table}

As shown in Table~\ref{tab:backbone_replacement}, every non-DINOv3 alternative backbone degrades performance relative to DINOv3 ViT-S/16 on Kvasir-SEG, and the no-branch variant drops to 86.23\%, confirming that the gains of the DFG branch come from the DINOv3 representation rather than the mere presence of an auxiliary path. The advantage is not explained by model scale alone, since the smaller self-supervised ViT-S/16 outperforms the larger CLIP ViT-B/16 and DINOv2 ViT-B/14, with a paired Wilcoxon test giving $p<0.05$; this suggests that the DINOv3 features are better suited to dense medical segmentation. Among the DINOv3 family, larger variants bring further gains, reaching 92.54\% Dice; we nevertheless adopt ViT-S/16 as the default for its accuracy-to-cost balance. A fully supervised ResNet-50 transfers the least. On ISIC2016 the margins are smaller, with the best non-DINOv3 alternative, MedSAM ViT-B, at 92.22\% against 92.58\% for ViT-S/16, but the ordering is preserved. The advantage is largest on the unseen PolypGen set, where ViT-S/16 attains 83.92\% Dice versus 78.45\% for the closest self-supervised alternative and 73.20\% for ResNet-50 (full PolypGen results for all backbones are reported in the supplementary material), indicating that the DINOv3 representation contributes most to cross-domain robustness.

\textbf{Hyperparameter Sensitivity.} We investigate the weight $\lambda$ of the auxiliary segmentation loss and the number of active DINOv3 feature layers injected into the denoising U-Net. Fig.~\ref{fig:hyperparam} reports the Dice scores on Kvasir-SEG for different configurations using the ViT-S/16 backbone.

Increasing the number of active DINOv3 layers consistently improves segmentation accuracy across all $\lambda$ values, with the 12-layer configuration achieving the highest Dice scores, confirming that richer hierarchical discriminative priors from deeper layers contribute positively to the denoising process. The gain from 6 to 12 layers is small relative to the added computational cost, so we retain 6 layers as the default. Regarding the loss weight, the model exhibits stable performance for $\lambda$ in $\{0.1, 0.3, 0.5, 0.7, 0.9\}$, with only marginal degradation at extreme values, indicating that the framework is not highly sensitive to this parameter. Considering the trade-off between accuracy and computational cost, we adopt $\lambda = 0.3$ with 6 active layers as the default configuration for the main experiments and all ablation studies, demonstrating that a moderate balance between the diffusion denoising objective and the auxiliary segmentation supervision yields good convergence.

\subsection{Color Sensitivity Analysis}
Because the instruction-driven formulation relies on color-coded editing targets, we assess the impact of color diversity during training. We build five training configurations with $N_c \in \{3, 5, 8, 12, 16\}$ colors, drawn in order from the fixed palette in Table~\ref{tab:supp_palette}; the main experiments adopt the $N_c=5$ configuration. All models are trained on the Kvasir-SEG dataset and evaluated with the instruction fixed to \emph{``Segment the polyp region using red,''} regardless of the color set used during training. Table~\ref{tab:supp_nc} reports the Dice and IoU on the Kvasir test set for each $N_c$. The results indicate that performance remains relatively stable across $N_c$, with a mild peak at $N_c=12$. Increasing color diversity therefore does not harm the editing capability when a requested color has been seen, and a moderate number of colors is sufficient for learning the instruction-to-color mapping.

\begin{table}[t]
\centering
\caption{Color palette used for instruction-driven training. The colors are arranged in the order used for constructing the training subsets with increasing $N_c$.}
\label{tab:supp_palette}
\renewcommand{\arraystretch}{1.1}
\begin{tabular}{@{}cll@{}}
\toprule
Index & Color Name & RGB Values \\ \midrule
1  & Red     & (255, 0, 0)   \\
2  & Green   & (0, 255, 0)   \\
3  & Blue    & (0, 0, 255)   \\
4  & Yellow  & (255, 255, 0) \\
5  & Purple  & (128, 0, 128) \\
6  & Pink    & (255, 192, 203) \\
7  & Cyan    & (0, 255, 255) \\
8  & Teal    & (0, 128, 128) \\
9  & Magenta & (255, 0, 255) \\
10 & Orange  & (255, 165, 0) \\
11 & Lime    & (50, 205, 50)  \\
12 & Coral   & (255, 127, 80) \\
13 & Violet  & (238, 130, 238) \\
14 & Navy    & (0, 0, 128)   \\
15 & Olive   & (128, 128, 0) \\
16 & Maroon  & (128, 0, 0)   \\ \bottomrule
\end{tabular}
\end{table}

\begin{table}[t]
\centering
\caption{Segmentation performance on Kvasir-SEG under different numbers of training colors ($N_c$). The inference instruction always specifies the color ``red''.}
\label{tab:supp_nc}
\renewcommand{\arraystretch}{1.2}
\begin{tabular}{@{}ccc@{}}
\toprule
$N_c$ & Dice (\%) & IoU (\%) \\ \midrule
3  & 91.25 & 86.43 \\
5  & 92.10 & 87.05 \\
8  & 91.28 & 86.38 \\
12 & 92.42 & 87.41 \\
16 & 91.49 & 86.46 \\ \bottomrule
\end{tabular}
\end{table}

\begin{figure}
\centering
\includegraphics[width=0.85\linewidth]{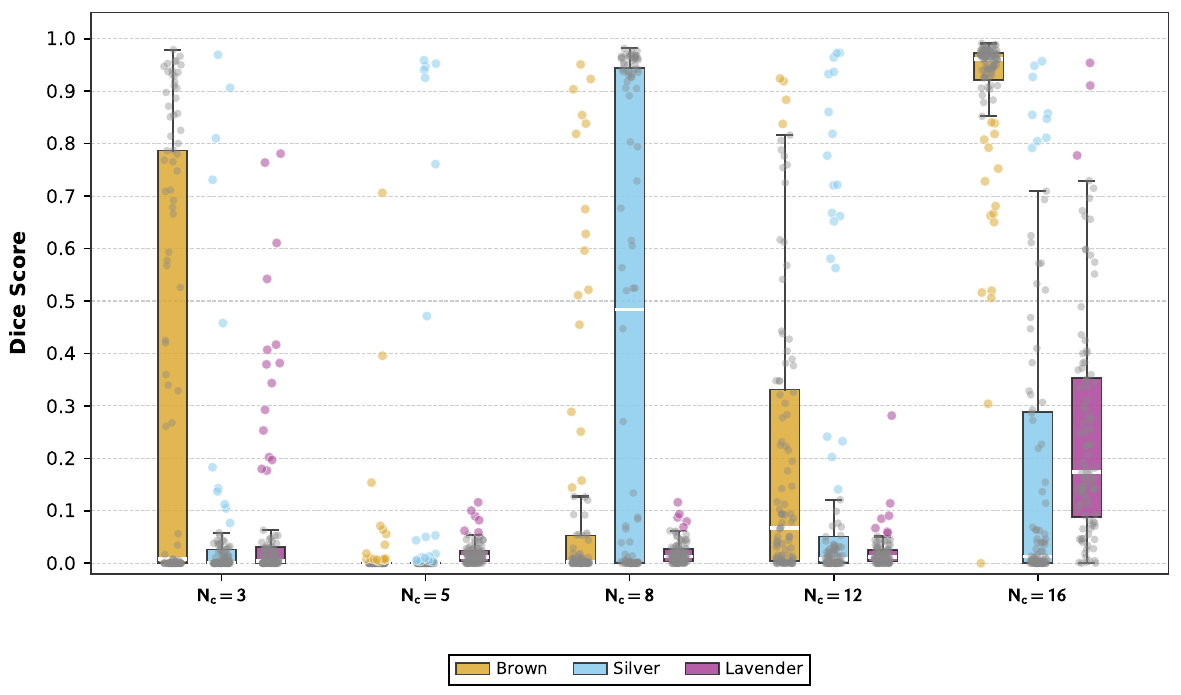}
\caption{Boxplots of Dice scores for three unseen colors.}
\label{fig:supp_boxplot}
\end{figure}

\subsubsection{Unseen Colors: Raw versus Corrected Evaluation}
Fig.~\ref{fig:supp_boxplot} shows the raw segmentation performance on three unseen colors, brown, silver, and lavender, none of which belongs to the 16-color training palette. The raw Dice conflates two distinct failure modes: the model may fail to segment the lesion at all, or it may segment correctly but paint with a wrong color, in which case the fixed color-map evaluation assigns every pixel to the wrong class and collapses the score. To separate these modes, we extract the dominant color actually painted by each model, remap the evaluation color map to this color, and recompute a corrected Dice. Table~\ref{tab:supp_decomp} reports raw and corrected Dice for all models and unseen colors.

\begin{table}[t]
\centering
\caption{Raw versus corrected Dice (\%) on unseen colors. Corrected Dice remaps the evaluation color map to the color actually painted by the model, isolating color-mapping artifacts from true segmentation failures.}
\label{tab:supp_decomp}
\renewcommand{\arraystretch}{1.15}
\begin{tabular}{@{}clcc@{}}
\toprule
$N_c$ & Unseen color & Raw Dice & Corrected Dice \\ \midrule
3 & brown & 26.66 & 35.13 \\
3 & silver & 3.81 & 60.23 \\
3 & lavender & 14.48 & 79.43 \\
5 & brown & 0.59 & 14.70 \\
5 & silver & 8.61 & 12.12 \\
5 & lavender & 2.20 & \textbf{92.72} \\
8 & brown & 17.46 & 32.28 \\
8 & silver & 26.28 & 51.33 \\
8 & lavender & 2.18 & \textbf{91.83} \\
12 & brown & 39.18 & 70.39 \\
12 & silver & 17.22 & 51.58 \\
12 & lavender & 2.03 & \textbf{92.37} \\
16 & brown & \textbf{92.13} & 92.55 \\
16 & silver & 26.96 & 68.93 \\
16 & lavender & 28.26 & \textbf{92.71} \\ \bottomrule
\end{tabular}
\end{table}

Three findings follow from Table~\ref{tab:supp_decomp}. First, the low raw Dice on lavender is almost entirely a color-mapping artifact, since the corrected Dice is above 90\% for all $N_c \geq 5$, meaning the model segments correctly but renders purple, the nearest color in its training pool, instead of lavender. Second, genuine segmentation failures on unseen colors persist at small $N_c$, with corrected Dice of 14.70\% and 12.12\% for brown and silver at $N_c=5$, but largely disappear at $N_c=16$, where brown reaches 92.55\%; training with a richer color pool decouples the color token from the task. Third, for lavender the apparent failure is essentially an evaluation artifact, whereas for brown and silver genuine segmentation failures coexist with color binding at small $N_c$; both issues largely disappear once the color pool is rich enough.

\begin{figure}
\centering
\includegraphics[width=0.95\linewidth]{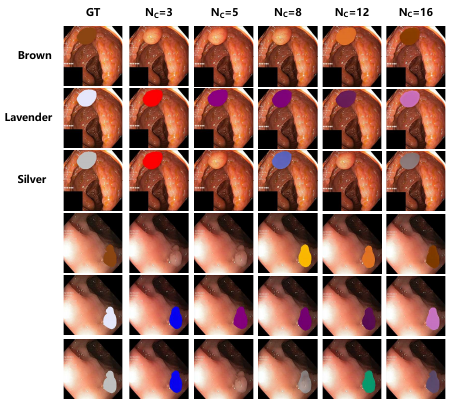}
\caption{Segmentation outputs on two test samples from models trained with different $N_c$. When the target color is unseen, models either reconstruct the original input directly or segment the polyp using a color that appeared during their training.}
\label{fig:supp_examples}
\end{figure}

Fig.~\ref{fig:supp_examples} visualizes the output on two test samples. When the target color is unseen, models either reconstruct the input without an overlay or segment the polyp using a training color. With $N_c=16$ the model produces an accurate brown mask, and for the harder silver and lavender requests it renders an incorrect but consistent color while preserving the segmentation, which suggests that a richer color set encourages more robust visual reasoning.

\section{Discussion}
\label{sec:discussion}
Our results support the editing formulation as a practical alternative to both discriminative models and binary-mask diffusion segmentation. Because the color-coded overlay retains the original image content while superimposing the target annotation, a pretrained latent diffusion model can operate on medical imagery with a much smaller effective domain shift. This is reflected in the unseen-domain results, where the framework attains the best Dice on both PolypGen and ISIC2017, and it explains why the overlay outperforms a standalone mask or a black-background mask in the output-representation ablation.

Two analyses bound the framework. The color-coded interface is robust to design choices, since performance stays within a narrow band across palettes and the unseen-color analysis shows that the model tends to render the nearest training color whenever segmentation succeeds; both color binding and genuine segmentation failures under unseen colors largely vanish with a sufficiently rich palette. In contrast, instruction control is limited to category selection, since the model does not parse spatial or size modifiers and attribute-conditioned selection is not supported.

These strengths come with trade-offs. In-domain, the strongest discriminative baselines remain competitive or superior on Kvasir-SEG, CVC-ClinicDB, and ISIC2016, and the editing formulation is more expensive at inference than a single discriminative pass, so its value is concentrated in cross-domain generalization, multi-lesion completeness, and instruction-conditioned control.

\section{Conclusion}
In this paper, we proposed InstEditSeg, a unified generative framework that recasts medical image segmentation as an instruction-driven diffusion editing task. By producing a color-coded edited image instead of a standalone binary mask, it differs from both binary-mask diffusion segmentation and discriminative decoders and aligns the output with the priors of latent diffusion models, reducing the domain gap between natural and medical images. A DINOv3-guided multi-scale feature pyramid, integrated through concatenation and zero-initialized convolution, injects hierarchical discriminative priors into the denoising process and improves lesion localization and boundary delineation. Experiments on polyp and skin lesion benchmarks show that the framework is competitive with strong discriminative baselines and attains the best Dice on the unseen PolypGen and ISIC2017 sets. The editing formulation further gives concrete advantages in multi-lesion completeness (12.75 Dice points over EMCAD), instruction-conditioned task control, and sampling flexibility, retaining about 99.7\% of the 25-step accuracy at five steps. The framework also has clear limitations: attribute-conditioned selection is unsupported, and the strongest discriminative baselines remain competitive or superior in-domain on Kvasir-SEG, CVC-ClinicDB, and ISIC2016. We plan to extend the framework to attribute-conditioned and interactive editing, few-step distillation, and 3D volumetric segmentation.

\section*{Declaration of generative AI and AI-assisted technologies in the manuscript preparation process}
During the preparation of this work the authors used DeepSeek in order to assist with LaTeX formatting. After using this tool, the authors reviewed and edited the content as needed and take full responsibility for the content of the published article.

\section*{Acknowledgment}
This work was supported by the National Natural Science Foundation of China under Grant 62572406, the Science and Technology Department of Sichuan Province under Grant 2024NSFSC2040.

{
    \small
    \bibliographystyle{ieeetr}
    \bibliography{main}
}
\end{document}